\documentclass[letterpaper, 10 pt, conference]{ieeeconf}  % Comment this line out if you need a4paper

\IEEEoverridecommandlockouts                              % This command is only needed if 
\usepackage{graphics}      % for pdf, bitmapped graphics files
\usepackage{epsfig}        % for postscript graphics files

\usepackage{newtxtext}    
\usepackage{newtxmath}    
\usepackage{adjustbox}
\usepackage{float} 
\usepackage{booktabs}     
\usepackage{subcaption}   
\usepackage{multirow}     
\usepackage{pifont}       
\usepackage{float}
\usepackage[misc]{ifsym}
\newcommand{\cmark}{\ding{51}} 
\newcommand{\xmark}{\ding{55}} 
\title{\LARGE \bf
Leveraging Geometric Priors for Unaligned Scene Change Detection
}

\author{Ziling Liu$^{1*}$, Ziwei Chen$^{1*}$, Mingqi Gao$^{3}$, Jinyu Yang$^{4}$, Feng Zheng$^{1,2,}$\textsuperscript{\Letter}
\thanks{$^{*}$Equal contribution.}%
\thanks{Ziling Liu and Ziwei Chen are with the Southern University of Science and Technology, Shenzhen, China (e-mail: \{liuzl2024, chenzw2023\}@sustech.edu.cn).}%
\thanks{Mingqi Gao is with the University of Sheffield, U.K. (e-mail: mingqi.gao@outlook.com).}%
\thanks{Jinyu Yang is with Tapall.ai (e-mail: jinyu.yang96@outlook.com).}%
\thanks{Feng Zheng is with the Southern University of Science and Technology, Shenzhen, China, and also with Spatialtemporal AI (e-mail: f.zheng@ieee.org).}%
\thanks{\textsuperscript{\Letter}Corresponding author.}%
}

\newcommand{\imgwidth}{0.12\textwidth}
\newcommand{\pscdwidth}{0.10\textwidth}
\begin{document}

\maketitle

\thispagestyle{empty}
\pagestyle{empty}

%%%%%%%%%%%%%%%%%%%%%%%%%%%%%%%%%%%%%%%%%%%%%%%%%%%%%%%%%%%%%%%%%%%%%%%%%%%%%%%%
\begin{abstract}

Unaligned Scene Change Detection aims to detect scene changes between image pairs captured at different times without assuming viewpoint alignment. To handle viewpoint variations, current methods rely solely on 2D visual cues to establish cross-image correspondence to assist change detection. However, large viewpoint changes can alter visual observations, causing appearance-based matching to drift or fail. Additionally, supervision limited to 2D change masks from small-scale SCD datasets restricts the learning of generalizable multi-view knowledge, making it difficult to reliably identify visual overlaps and handle occlusions. This lack of explicit geometric reasoning represents a critical yet overlooked limitation. In this work, we introduce geometric priors for the first time to address the core challenges of unaligned SCD, for reliable identification of visual overlaps, robust correspondence establishment, and explicit occlusion detection. Building on these priors, we propose a training-free framework that integrates them with the powerful representations of a visual foundation model to enable reliable change detection under viewpoint misalignment. Through extensive evaluation on the PSCD, ChangeSim, and PASLCD datasets, we demonstrate that our approach achieves superior and robust performance. 
Our code will be released at https://github.com/ZilingLiu/GeoSCD.

\end{abstract}

\section{INTRODUCTION}

Scene Change Detection (SCD) has broad applicability across multiple domains, including scene updating~\cite{clsplats, 3dgs-cd, lt-gs}, object-level navigation in robotic systems~\cite{gs-lts}, and environmental perception in autonomous driving~\cite{vl_cmu_cd, pscd, tanet}.
While most existing SCD methods~\cite{tanet, changenet, transcd} are developed and evaluated under the assumption of viewpoint alignment, this constraint limits their applicability to real-world scenarios. Usually, images captured at different times often exhibit viewpoint discrepancies caused by sensor placement, motion, or environmental factors, as commonly encountered in autonomous driving, drone, or mobile robot applications. 

Several recent works have attempted to address unaligned viewpoints in SCD by establishing correspondences between image pairs, typically relying on 2D cues such as optical flow~\cite{dof-cdnet}, feature correlation~\cite{pscd}, or attention-based semantic alignment~\cite{rbcd}, and then integrating these correspondences with a change decoder to predict changes.
Although these methods achieve reasonable performance on existing evaluation datasets~\cite{vl_cmu_cd, pscd} with limited and homogeneous viewpoint variations, they struggle when applied to more complex scenes with diverse viewpoint changes. This is mainly because these approaches typically rely on supervision from only 2D change masks and are trained on limited SCD data. As a result, they fail to learn generalizable multi-view knowledge, hindering their ability to maintain robust correspondence and performance across diverse viewpoint changes. Furthermore, relying solely on 2D information presents significant challenges when dealing with unaligned viewpoints. For example, these methods often struggle to robustly determine the visual overlap between images and exclude irrelevant information outside the overlapping regions. Additionally, viewpoint shifts inevitably cause occlusions, where regions visible in one image are obscured in the other. With only 2D cues, it is difficult to explicitly detect occluded areas and avoid misinterpreting them as changes. In contrast, 3D information provides a natural solution to these challenges. Just as humans can effortlessly establish associations and perceive occlusions when observing a scene from different viewpoints thanks to our innate geometric reasoning, robust unaligned SCD also requires explicit geometric understanding.

Recently, pretrained geometric foundation models (GFMs)~\cite{dust3r, vggt, fast3r, cut3r} have shown strong generalization in recovering 3D geometry, such as poses, depth, and structure, directly from multi-view images across diverse scenes and viewpoint disparities.
\begin{figure}[!t]
    \centering
    \includegraphics[width=0.48\textwidth]{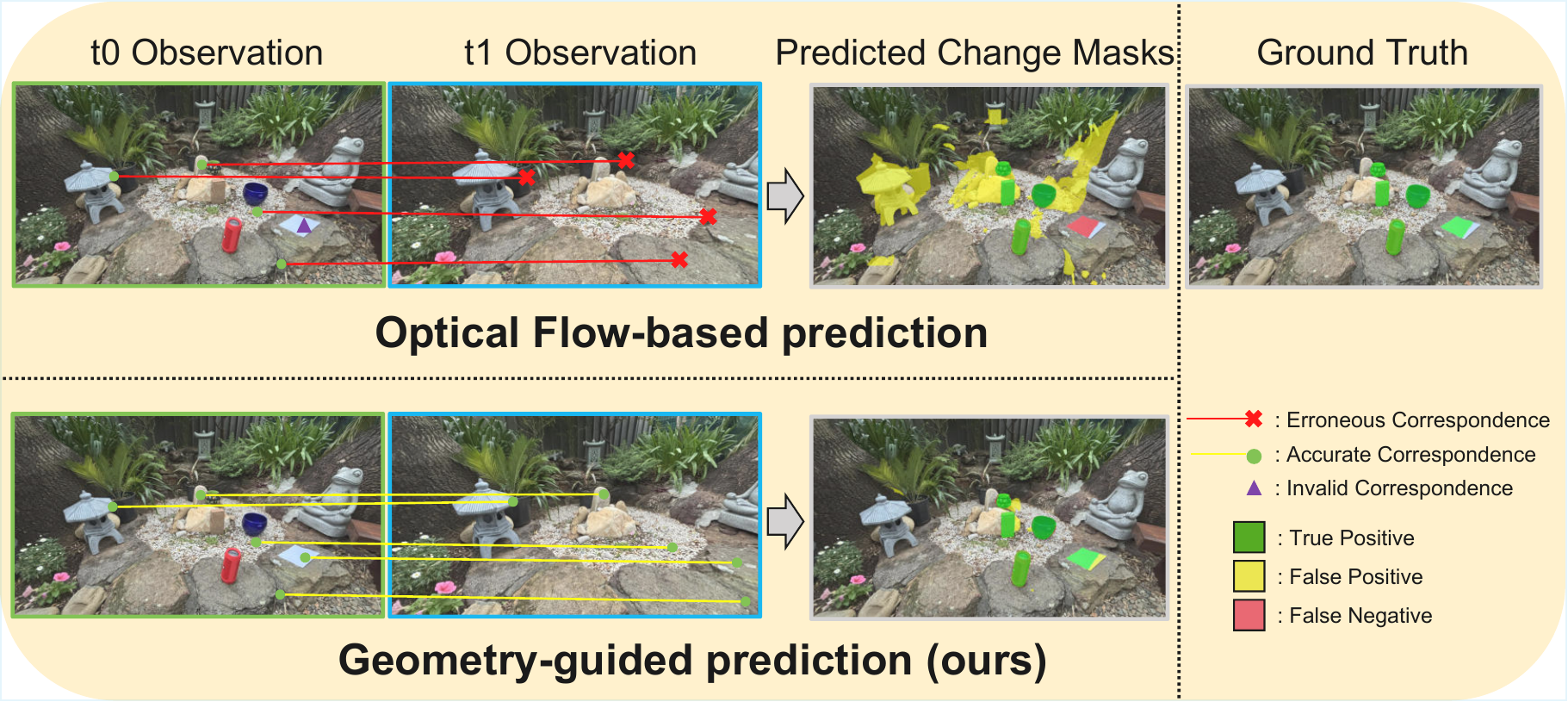}
    \caption{Comparison between GFM-based geometric correspondences and optical flow-based correspondences. Invalid correspondence means that the matched point falls outside the image bounds. Leveraging geometric correspondences, our method generates more accurate change masks.}
    \label{fig:motivation}
\end{figure}
In real-world scenarios, when scene changes occur, the global geometric structure of the scene typically remains consistent. Under such conditions, GFMs can robustly reconstruct per-pixel geometry and camera poses from images captured at different time points.
This enables a unified 3D representation of the scene that is invariant to viewpoint changes, providing consistent holistic geometry before and after changes. By reprojecting pixels using estimated camera poses, we can effectively identify overlapping regions and establish robust correspondences before and after changes. These projection-based geometric correspondences also naturally handle regions that have undergone changes, outperforming 2D feature-based methods that rely on appearance consistency. We demonstrate the advantage of geometric correspondence over visual correspondence for unaligned SCD in Fig.~\ref{fig:motivation}.
Moreover, GFM reconstruction enables explicit detection of occluded regions, allowing us to effectively exclude them from genuine change areas. 

Based on these capabilities, we propose a novel two-stage framework that first leverages GFM to establish robust geometric understanding and then utilizes it to guide change mask prediction. This approach integrates the power of the visual foundation model SAM~\cite{sam}, drawing inspiration from zero-shot methods~\cite{gescf,zero} to achieve strong cross-dataset generalization.
We evaluated our method against several state-of-the-art works on the PSCD, ChangeSim, and PASLCD datasets. The results demonstrate that our approach consistently maintains superior and robust performance across diverse viewpoint changes and scenes. Our contributions are as follows: 

\begin{itemize}
    \item We are the first to introduce geometric priors from GFM to accurately identify visual overlaps, establish robust correspondences, and explicitly detect occlusions, which are difficult to achieve with 2D visual cues only.

    \item We propose a novel training-free framework that integrates geometric priors with a visual foundation model, eliminating the need for large-scale annotated datasets and enhancing generalization capability.  
    
    \item Through extensive evaluation on PSCD, ChangeSim, and PASLCD datasets, we demonstrate that our method achieves leading and robust performance under various viewpoint changes and scene types. Our ablation studies further validate the effectiveness of geometric priors for unaligned SCD.

\end{itemize}

\section{RELATED WORK}
\subsection{Scene Change Detection}

With the advance of deep feature learning, numerous SCD methods based on deep neural networks have emerged. In traditional aligned settings, mainstream approaches~\cite{changenet, c3po, transcd, tanet} typically adopt Siamese CNN~\cite{siamese} or Transformer~\cite{transformer} architectures with multi-scale feature fusion or temporal attention to improve feature alignment and change modeling, achieving strong performance on aligned datasets. Meanwhile, with the rise of visual foundation models, some studies~\cite{gescf, zero} have also explored training-free SCD frameworks, such as leveraging tracking~\cite{deva} or segmentation models~\cite{sam} to perform zero-shot change detection, thereby avoiding the need for large-scale annotated datasets.
For the more challenging case of unaligned viewpoints, several works have made initial attempts. For example, DOF-CDNet~\cite{dof-cdnet} employs optical flow to align cross-temporal image pairs, but it is prone to errors under large viewpoint differences or in the presence of object motion. CSCDNet~\cite{pscd} introduces a correlation layer in the decoder to enhance robustness under large viewpoint shifts. Despite this, the method remains limited by its reliance on feature correlation, which lacks explicit geometric reasoning and hinders generalization in complex 3D scenes. Furthermore, its local window-based correlation strategy proves inadequate for handling significant viewpoint changes.
In contrast, RSCD~\cite{rbcd} utilizes frozen DINOv2~\cite{dinov2} features and bidirectional cross-attention to establish correspondences. Nevertheless, without geometric grounding in its backbone and due to its dependence on semantic alignment, the approach struggles to generalize across diverse multi-view data and is highly susceptible to geometric variations in the scene, such as spatial layout shifts and viewpoint disparities.

\subsection{Geometric Foundation Model}

Geometric foundation models (GFMs) are attention-based models that directly infer core 3D properties including camera parameters, depth, and per-pixel point maps from images, thereby providing explicit multi-view geometric priors~\cite{dust3r}. Recent advances~\cite{fast3r,vggt,cut3r, flare} have shifted the focus from robust two-view point-map regression~\cite{dust3r, mast3r} toward scalable multi-view and multi-attribute prediction. Beyond reshaping traditional reconstruction pipelines, GFMs have demonstrated strong impact across diverse downstream tasks. For example, they enable realistic scene and camera control in video generation~\cite{viewcrafter}, replace traditional pose optimization with direct point-map regression in SLAM~\cite{slam3r}, and enhance representation learning for 3D scene understanding~\cite{spatialmllm}. These applications highlight the effectiveness of GFMs’ geometric priors for processing multi-view data and boosting downstream performance. Motivated by these advances, we are the first to adopt GFMs in unaligned SCD to address the key challenges of viewpoint misalignment

\section{Method}

\subsection{Overview}

\begin{figure*}[t]
    \centering
    \includegraphics[width=0.95\textwidth]{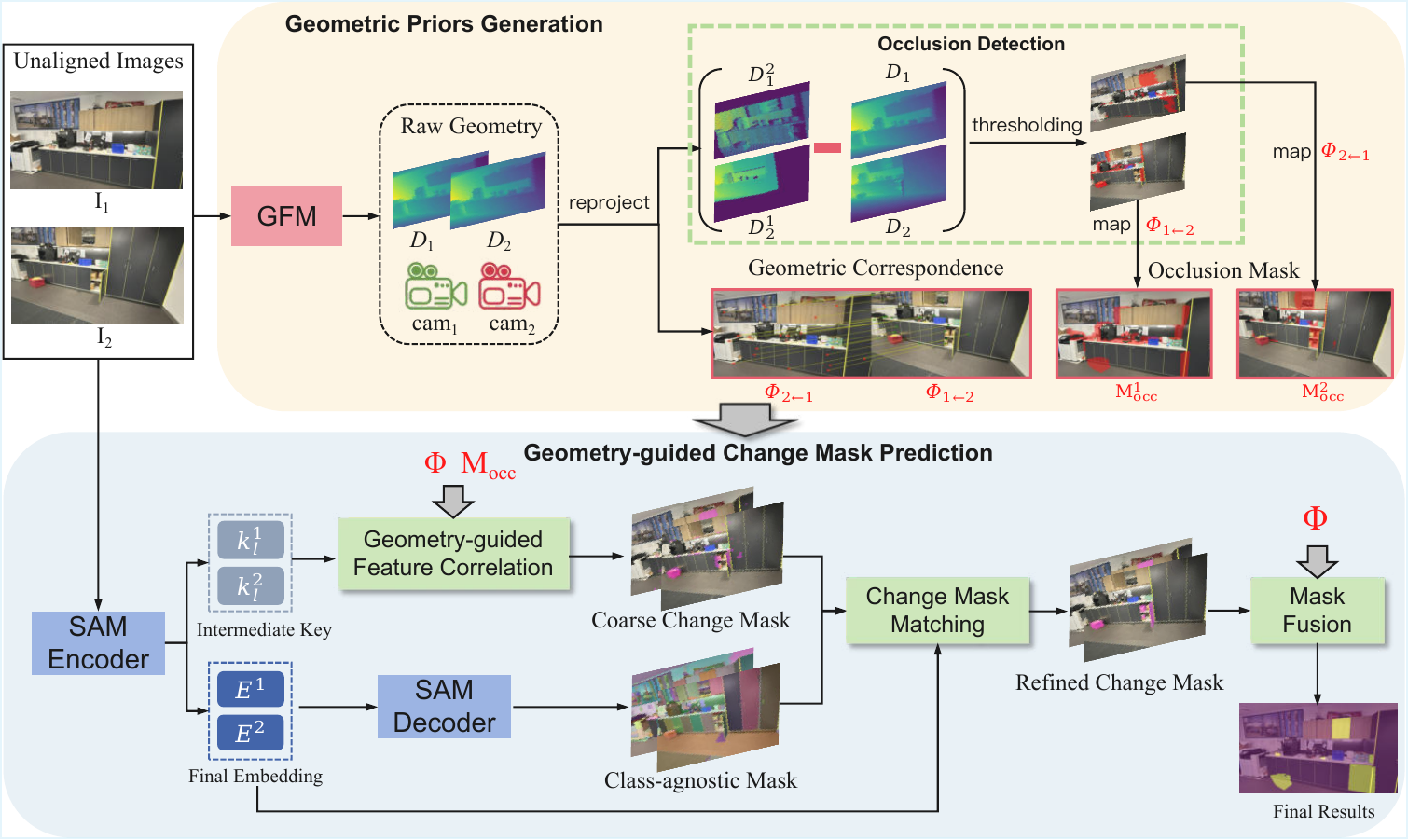}
    \caption{Framework overview. We first derive geometric correspondences, visual overlaps, and occlusion masks from GFM reconstructions. These cues then guide SAM to generate coarse change masks via feature correlation, which are further refined and fused to produce the final results.}
    \label{fig:architecture}
    % \vspace{-1em}
\end{figure*}

As shown in Fig.~\ref{fig:architecture}, our framework comprises two core modules: the Geometric Priors Generation module and the Geometry-guided Change Mask Prediction module.
Given an image pair $\{I_1, I_2\}$ captured at two different times with unaligned viewpoints, where $I_1$ serves as the query image and $I_2$ serves as the reference image, the Geometric Priors Generation Module leverages the GFM to recover depth maps and camera parameters. In this work, we adopt VGGT~\cite{vggt} due to its demonstrated accuracy and efficiency. Using these reconstruction results, the module establishes pixel-wise correspondences and identifies the visual overlap between the image pair. Furthermore, it explicitly detects occluded regions to allow subsequent filtering.
The Geometry-guided Change Mask Prediction module then integrates these geometric cues with the Visual Foundation Model SAM~\cite{sam}, inspired by zero-shot SCD approaches~\cite{mv3dcd}. Guided by the geometric correspondences, initial change proposals are generated within overlap regions by measuring the similarity between the SAM encoder intermediate features of matched pixels. These proposals are then refined using the occlusion masks to produce coarse change masks.
Subsequently, the coarse masks are matched with class-agnostic segmentation masks from SAM based on mask intersection and semantic consistency, yielding refined change masks for $I_1$ and $I_2$. Finally, the two masks are fused using the geometric correspondences to produce the final result. 
Additionally, we introduce a preprocessing step that mitigates the impact of significant illumination changes before feeding images into the GFM.

\subsection{Geometric Priors Generation}
\paragraph{Geometric Correspondence Generation}
We adopt the Geometric Foundation Model (GFM) $\mathscr{G}$ to reconstruct geometric information from the input image pair $\{I_1, I_2\}$:
\begin{equation}
\mathcal{K}, \mathcal{T}, \mathcal{D} = \mathscr{G}(I_1, I_2),
\end{equation}
where $\mathcal{K} = \{K_1, K_2\}$ denotes the camera intrinsics, $\mathcal{T} = \{T_1, T_2\}$ with $T_i \in \text{SE}(3)$ denotes the camera extrinsics, and $\mathcal{D} = \{D_1, D_2\}$ denotes the predicted depth maps. Let $\Omega_i = [0, W_i) \times [0, H_i)$ denote the pixel domain of image $I_i$, with $(W_i, H_i)$ being its width and height. For each pixel $p_1 \in \Omega_1$, we establish its correspondence $\Phi_{2\leftarrow 1}(p_1)\in \Omega_2$ by reprojecting $p_1$ using the estimated depth and camera parameters:
\begin{align}
x_2^1&=R_{2\leftarrow 1} \, D_1(p_1) \, K_1^{-1} \tilde{p}_1+t_{2\leftarrow 1} \label{eq:x2},\\
p_2^1&=\Phi_{2\leftarrow 1}(p_1) = \pi (K_2x_2^1 ) \label{eq:p2},
\end{align}
% We assume that $I_1$ is the query image at $t_0$ and $I_2$ is the reference image at $t_1$.
where $\tilde{p}_1$ is the homogeneous representation of $p_1$, $R_{2\leftarrow 1}$ and $t_{2\leftarrow 1}$ are the rotation and the translation in the transformation $T_{j\leftarrow i}=T_j T_i^{-1}=\begin{bmatrix}R_{j\leftarrow i}&t_{j\leftarrow i}\\0&1\end{bmatrix}$, $\pi$ is perspective projection and $p_2^1$ is the corresponding pixel of $p_1$ in $I_2$. Similarly, the inverse correspondence $\Phi_{1 \leftarrow 2}$ can be derived. We then determine the overlapping region in $I_1$ as the set of pixels whose reprojections fall within the valid image domain:
\begin{equation}
\Omega_{1 \rightarrow 2} = \left\{ p_1 \in \Omega_1 \mid p_2^1 \in \Omega_2  \right\} \label{eq:4}.
\end{equation}
Symmetrically, we can obtain the overlapping $\Omega_{2 \rightarrow 1}\subset \Omega_2$.

\begin{figure}[t]   % 不要 figure*，改成 figure
    
    \centering
    \includegraphics[width=\linewidth]{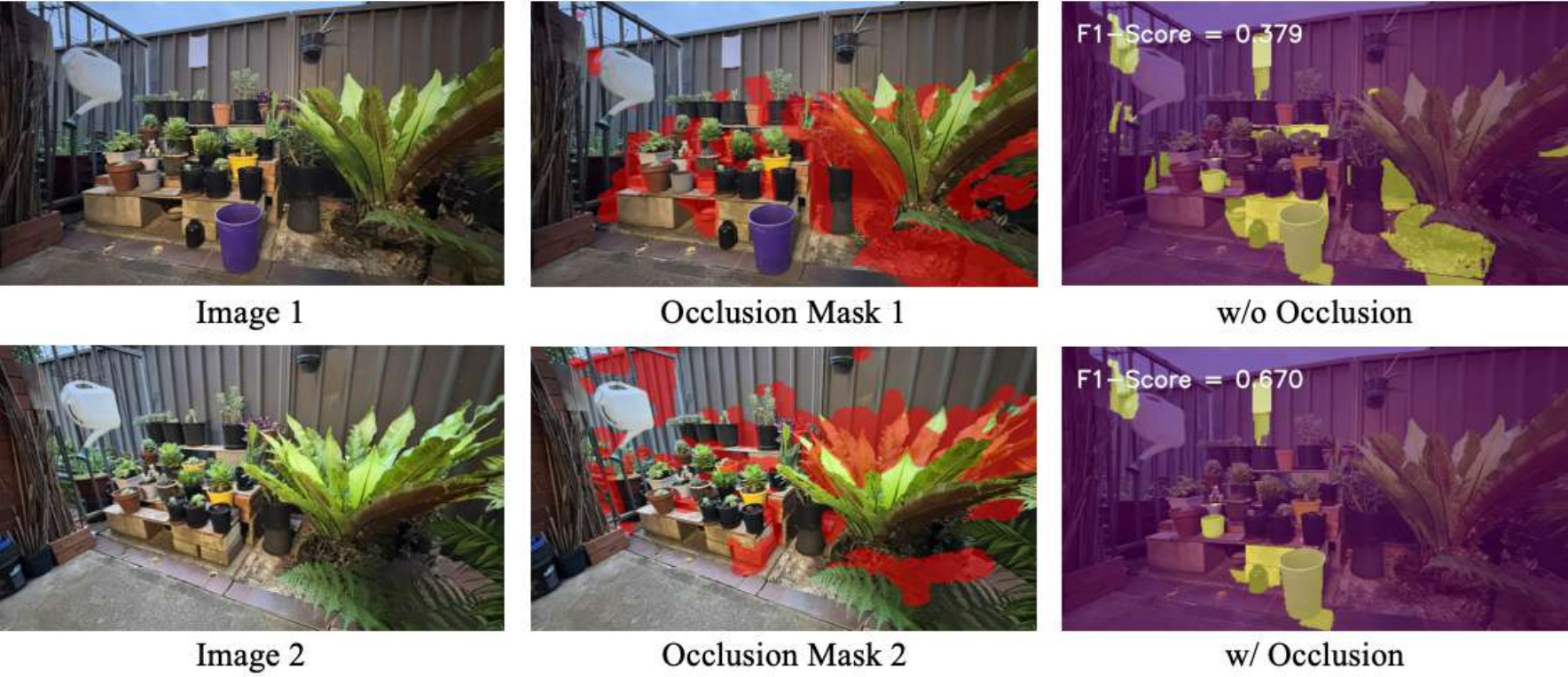}
    \caption{Visualization of detected occlusions. Red regions indicate occluded areas.}
    \label{fig:occupy}
\end{figure}

\paragraph{Occlusion Detection}
In addition to pixel-level correspondences, geometric information enables the identification of occluded regions across views. Specifically, for each pixel $p_1 \in \Omega_{1 \rightarrow 2}$ within overlap region in $I_1$, we compute its reprojected depth $D_2^1(p_1)$ at the corresponding location $p_2^1$ in $I_2$ using the 3D point $x_2^1$ obtained from Eq~\ref{eq:x2}:
\begin{equation}
D_2^1(p_1) = e_3^\top x_2^1,e_3=[0,0,1]^\top.
\end{equation}
The occlusion state of $p_1$ is then determined by comparing the reprojected depth with the estimated depth $D_2$ at the corresponding pixel in $I_2$:
\begin{equation}
    M_{\text{occ}}^1(p_1) = 
    \begin{cases}
        1, & \text{if } D_2^1(p_1) - D_2(p_2^1) > \tau, \\
        0, & \text{otherwise},
    \end{cases}  \label{eq:6}
\end{equation}
where $\tau$ is a depth consistency threshold. To ensure robustness across varying scene scales and depth uncertainties, $\tau$ is adaptively computed. It combines a geometric constraint proportional to the median scene depth and a statistical constraint based on the median absolute deviation (MAD) of the depth differences:
\begin{equation}
    \tau = \max\big(\alpha \cdot \mathrm{med}(D_2),\, \mathrm{med}(\Delta D) + \kappa \cdot \mathrm{MAD}(\Delta D)\big),
\end{equation}
where $\Delta D$ is the depth difference in Eq~\ref{eq:6},  $\alpha=0.03$, and $\kappa=2.5$.
Pixels that satisfy the above condition are visible in $I_1$ but occluded in $I_2$. 
Similarly, we can obtain $M_{\text{occ}}^2$. We visualize a pair of predicted occlusion masks in Fig.~\ref{fig:occupy}.

\subsection{Geometry-guided Change Mask Prediction}
\label{sec:geom_corr}

\paragraph{Geometry-guided feature correlation}
We feed the image pair $I_1$ and $I_2$ into the SAM encoder and extract multi-head key features from its $l$-th transformer layer, denoted as $k_l^1$ and $k_l^2$, with $l=17$ following ~\cite{gescf}. In addition, we also obtain the final-layer embedding of the encoder $E^1$ and $E^2$. All extracted features are then resized to image resolution for subsequent pixel-level operations.
For each pixel $p_1 \in \Omega_{1 \rightarrow 2}$ within overlap, we compute the cosine similarity between $k_l^1(p_1)$ and its geometrically corresponding feature $k_l^2(p_2^1)$ across all attention heads. The similarities from different heads are averaged to produce the similarity map $S_l^1$. An initial change proposal $P_{\text{init}}^1$ is derived by applying the Adaptive Threshold Function~\cite{gescf} to $S_l^1$.
To filter out occluded regions, we refine $P_{\text{init}}^1$ using:
\begin{equation}
P_{\text{ref}}^1 = P_{\text{init}}^1 \odot (1 - M_{\text{occ}}^1),
\end{equation}
where $M_{\text{occ}}^1$ denote the occlusion masks. The refined proposal $P_{\text{ref}}$ also serves as the coarse change mask.

\paragraph{Change Mask Matching \& Mask Fusion}
We then obtain the fine change mask in $I_1$ via Geometric-Semantic Mask Matching~\cite{gescf}, which selects class-agnostic SAM masks $\{\text{mask}_i\}_{\text{SAM}}$ that sufficiently overlap with $P_{\text{ref}}^1$ and exhibit semantic inconsistency across the two images:
\begin{equation}
M_{\text{change}}^1 = \text{GSM\_Match}\big(P_{\text{ref}}^1, \{\text{mask}_i\}_{\text{SAM}}\big).
\end{equation}
The same procedure is applied symmetrically to $I_2$, obtaining $M_{\text{change}}^2$. Using the established pixel correspondences $\Phi_{1 \leftarrow 2}$ between $I_1$ and $I_2$, we map $M_{\text{change}}^2$ onto the query view and fuse it with $M_{\text{change}}^1$ to obtain the final change mask:
\begin{equation}
M_{\text{final}} = M_{\text{change}}^1 \cup \Phi_{1 \leftarrow 2}(M_{\text{change}}^2).
\end{equation}

% 光照
\subsection{Handling Illumination Variations}

\begin{figure}[t]   % 不要 figure*，改成 figure
    \centering
    \includegraphics[width=\linewidth]{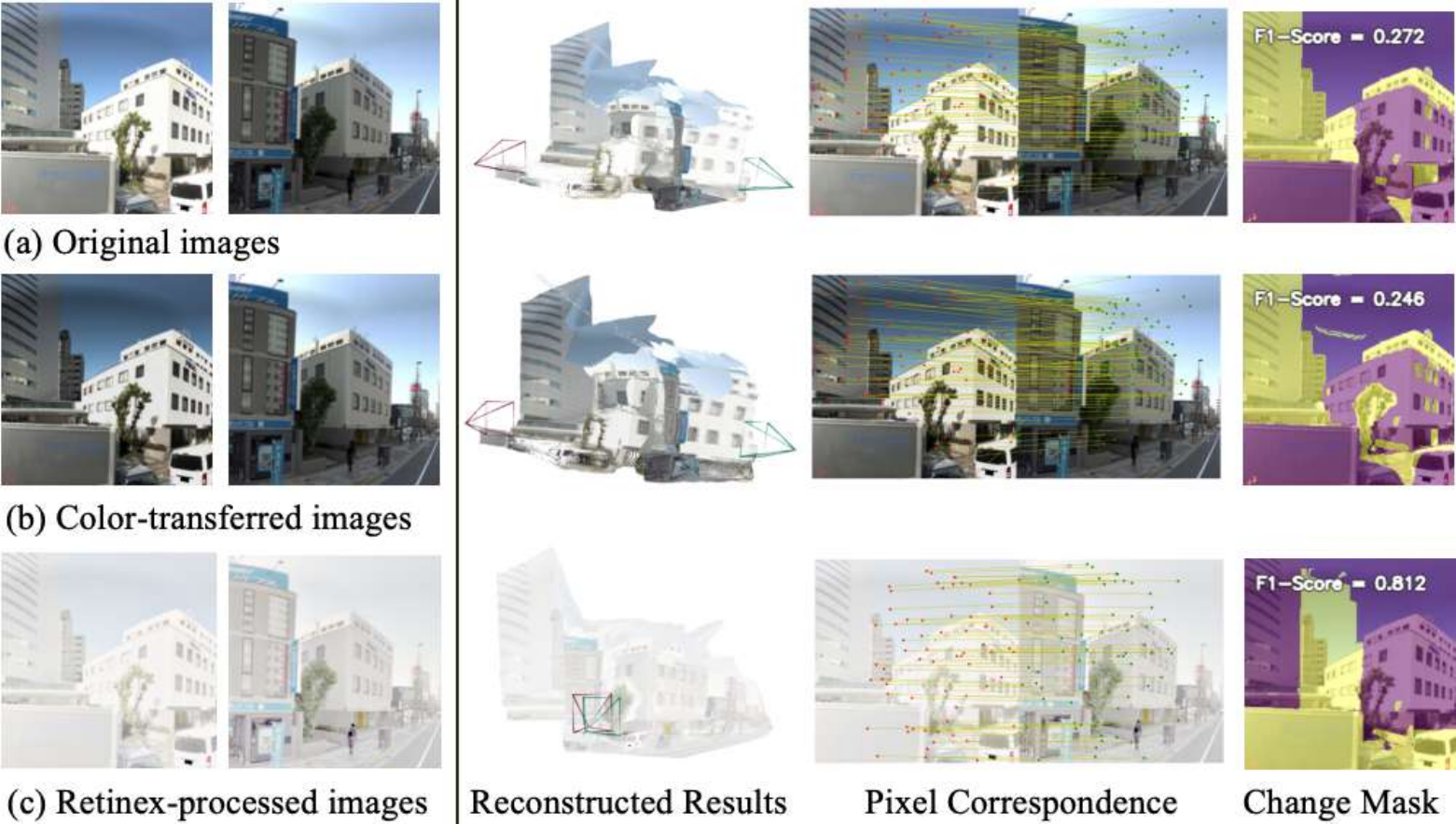}
    \caption{Effect of preprocessing on GFM reconstruction and change detection. Retinex improves robustness under illumination changes.}
    \label{fig:demo}
\end{figure}

Significant illumination or color differences between input views can cause GFM reconstruction failures, as shown in Fig.~\ref{fig:demo}(a). To address this, we apply preprocessing only before feeding images into GFM, when the brightness or color distribution difference between the input image pair exceeds a predefined threshold. We experiment with two approaches: Color Transfer~\cite{colortransfer} and Retinex~\cite{retinex}. Specifically, Retinex extracts the reflectance components of the image pair to suppress illumination variations, while Color Transfer attempts to reduce the RGB distribution gap between the images. Although Retinex produces images that differ significantly in appearance from the originals, it enforces stronger consistency between the input pair. Our experiments (Fig.~\ref{fig:demo}) demonstrate that GFM benefit from Retinex preprocessing, achieving more reliable reconstruction and improved performance under significant illumination changes.

\section {Experiments}

\subsection{Datasets}

\begin{table}[!t]
\centering
\caption{Properties of datasets after processing for unaligned SCD evaluation. $\dagger$ indicates the dataset is processed into unaligned settings.}
\label{tab:benchmark-summary}
\begin{tabular}{l|c|c|c}
\toprule
 & PSCD$^\dagger$~\cite{pscd} & ChangeSim$^\dagger$~\cite{changesim} & PASLCD$^\dagger$~\cite{mv3dcd} \\
\midrule
Image pairs   & 2079 & 4887 & 1500 \\
Data source   & Real  & Synthetic & Real \\
Resolution    & 224$\times$224 & 640$\times$480 & 504$\times$284 \\
Environment   & Outdoor & Indoor & Both \\
Camera intervals & stride-1/2 & stride-5/10/15 & stride-5/10/15 \\
Camera translation        & \xmark & \cmark & \cmark \\
Camera rotation       & \cmark & \cmark & \cmark \\
Rotation Range  & $\{10^{\circ}, 20^{\circ}\}$ & $(0^{\circ}, 90^{\circ}]$ & $(0^{\circ}, 90^{\circ}]$ \\
Aligned pair  & \cmark & \cmark & \xmark \\
\bottomrule
\end{tabular}
\end{table}

\begin{figure*}[!ht]
    \centering
    \setlength{\tabcolsep}{0pt}
    \renewcommand{\arraystretch}{0.5}
    \begin{tabular}{c @{\hspace{2pt}} c @{\hspace{4pt}} | @{\hspace{4pt}} c @{\hspace{1pt}} c @{\hspace{1pt}} c @{\hspace{1pt}} c @{\hspace{1pt}} c} 
        % 顶部表头
        $t_0$ image & $t_1$ image & CSCDNet~\cite{pscd} & RSCD~\cite{rbcd} & GeSCF~\cite{gescf} & Ours & GT \\ 
        \toprule

        % -------- 第一组 --------
        % \multirow{3}{*}{\includegraphics[width=\imgwidth]{figures/pal_7260/t0.jpg}}
        % \multirow{3}{*}{\rotatebox[origin=c]{90}{\textbf{PASLCD}}}
       % \multirow{3}{*}{\rotatebox[origin=c]{90}{\hspace{2cm} \textbf{PASLCD}}}
%        \multirow{3}{*}{
%   \makebox[2cm][c]{\hspace{-0.5cm}\rotatebox[origin=c]{90}{PASLCD~\cite{mv3dcd}}}
% }
        \phantom{\includegraphics[width=\imgwidth]{figures/pal_7260/t0.jpg}} &
        \includegraphics[width=\imgwidth]{figures/pal_7260/stride5_t1.jpg} &
        \includegraphics[width=\imgwidth]{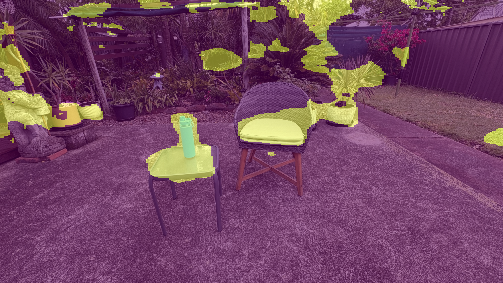} &
        \includegraphics[width=\imgwidth]{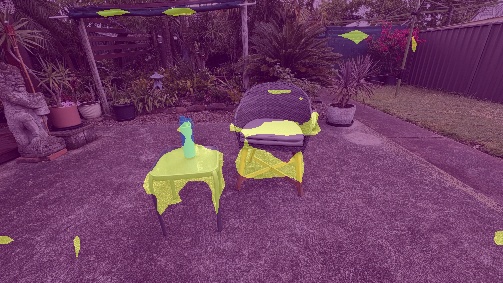} &
        \includegraphics[width=\imgwidth]{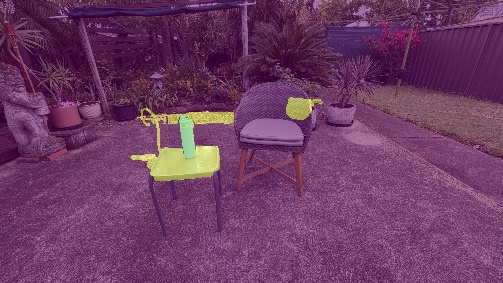} &
        \includegraphics[width=\imgwidth]{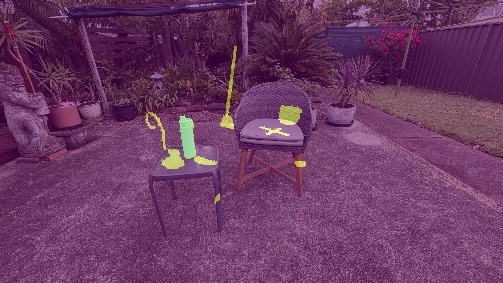} &
        \includegraphics[width=\imgwidth]{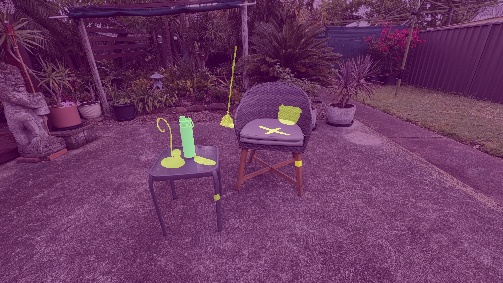}  \\

        \includegraphics[width=\imgwidth]{figures/pal_7260/t0.jpg}
        & \includegraphics[width=\imgwidth]{figures/pal_7260/stride10_t1.jpg} &
        \includegraphics[width=\imgwidth]{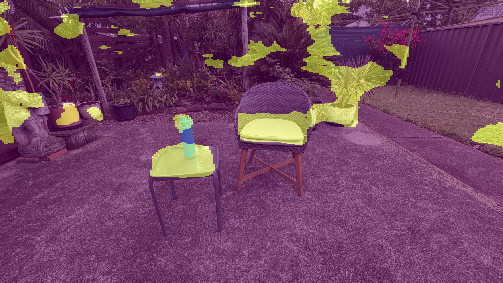} &
        \includegraphics[width=\imgwidth]{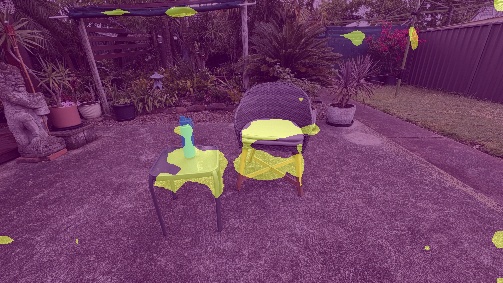} &
        \includegraphics[width=\imgwidth]{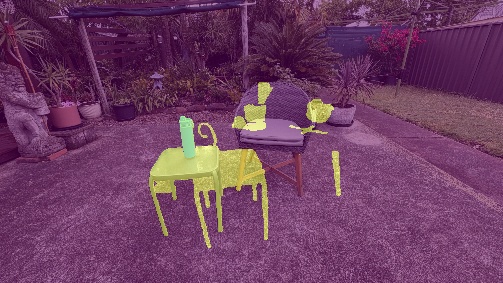} &
        \includegraphics[width=\imgwidth]{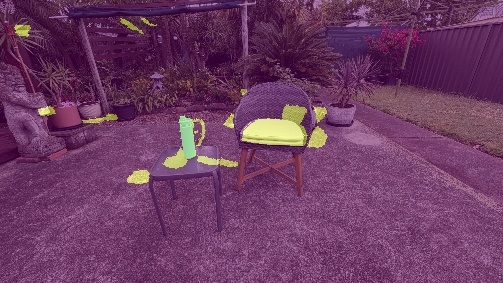} &
        \includegraphics[width=\imgwidth]{figures/pal_7260/gt.jpg}  \\

         \phantom{\includegraphics[width=\imgwidth]{figures/pal_7260/t0.jpg}} & \includegraphics[width=\imgwidth]{figures/pal_7260/stride15_t1.jpg} &
        \includegraphics[width=\imgwidth]{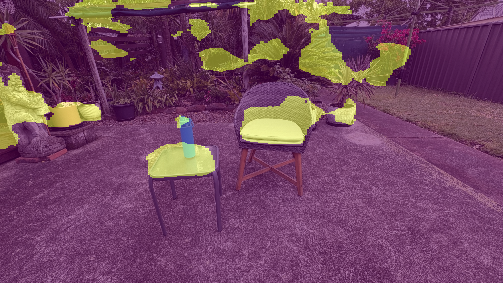} &
        \includegraphics[width=\imgwidth]{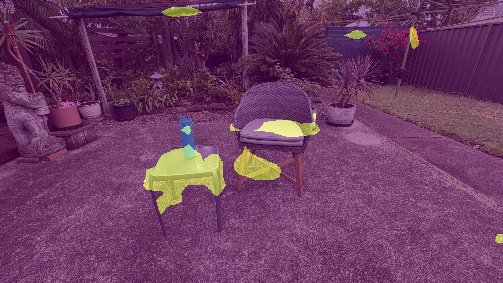} &
        \includegraphics[width=\imgwidth]{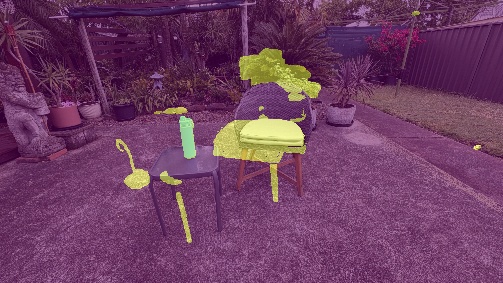} &
        \includegraphics[width=\imgwidth]{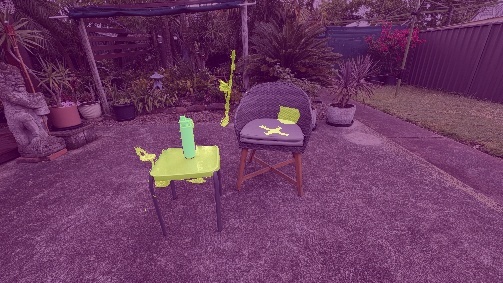} &
        \includegraphics[width=\imgwidth]{figures/pal_7260/gt.jpg}  \\

        \midrule

         % \multirow{3}{*}{\includegraphics[width=\imgwidth]{figures/05_use/t0.jpg}} 
         % \multirow{3}{*}{\rotatebox[origin=c]{90}{\textbf{PASLCD}}}
         \phantom{\includegraphics[width=\imgwidth]{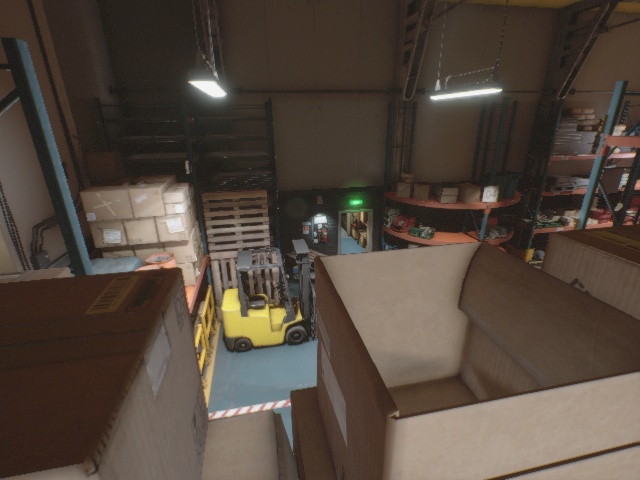}}&
        \includegraphics[width=\imgwidth]{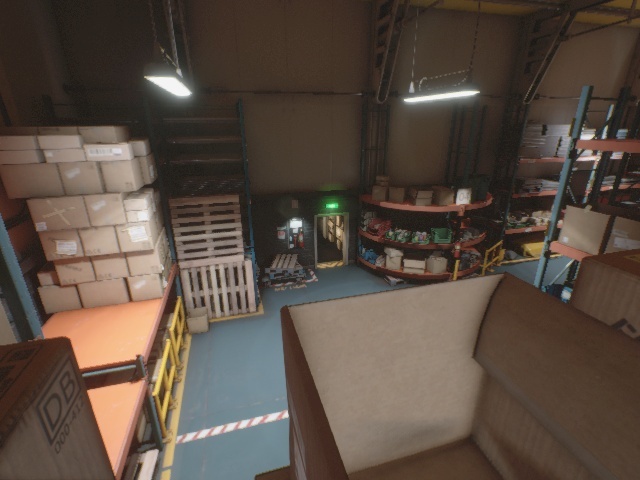} &
        \includegraphics[width=\imgwidth]{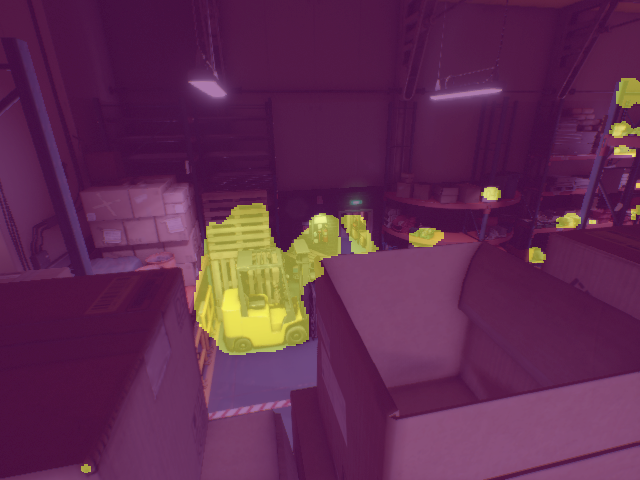} &
        \includegraphics[width=\imgwidth]{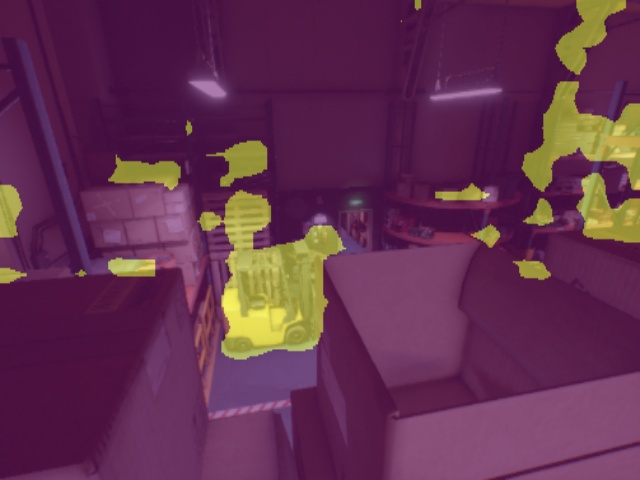} &
        \includegraphics[width=\imgwidth]{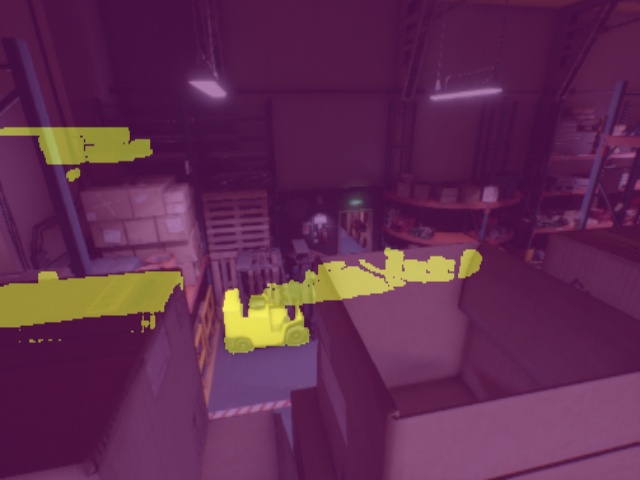} &
        \includegraphics[width=\imgwidth]{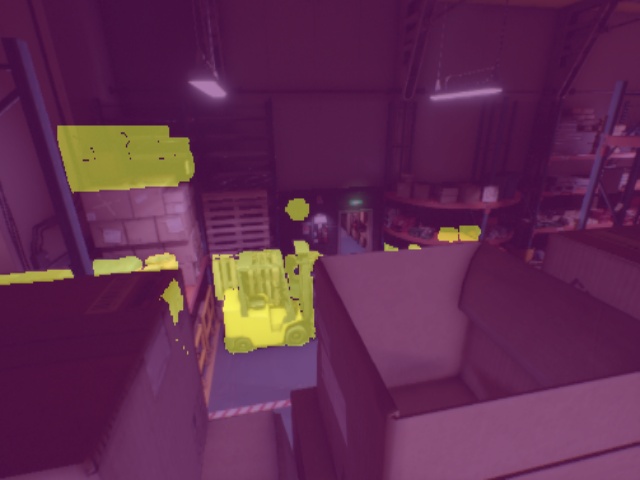} &
        \includegraphics[width=\imgwidth]{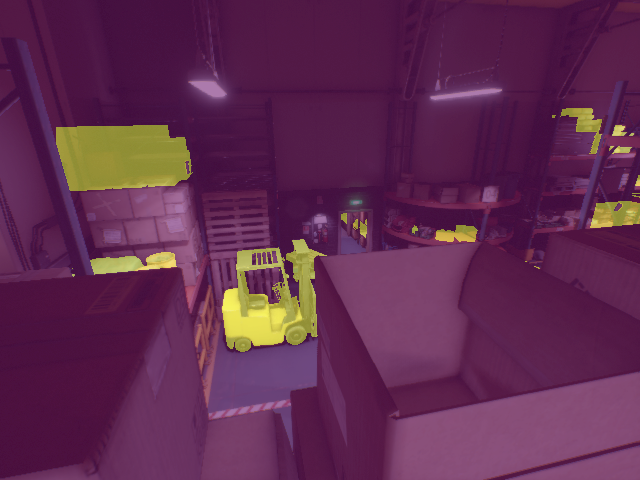}  \\

        \includegraphics[width=\imgwidth]{figures/05_use/t0.jpg} & \includegraphics[width=\imgwidth]{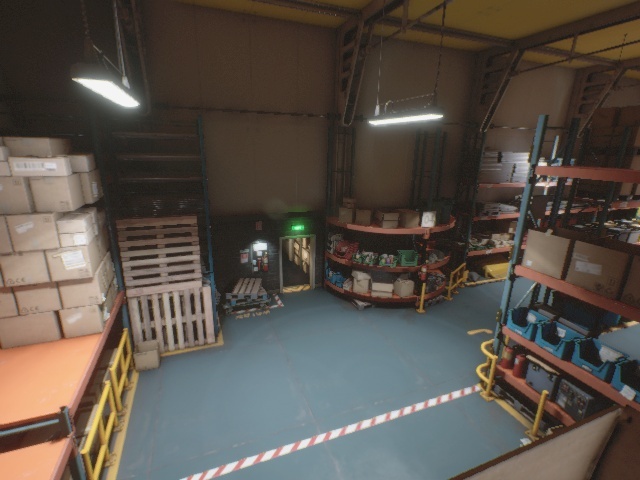} &
       \includegraphics[width=\imgwidth]{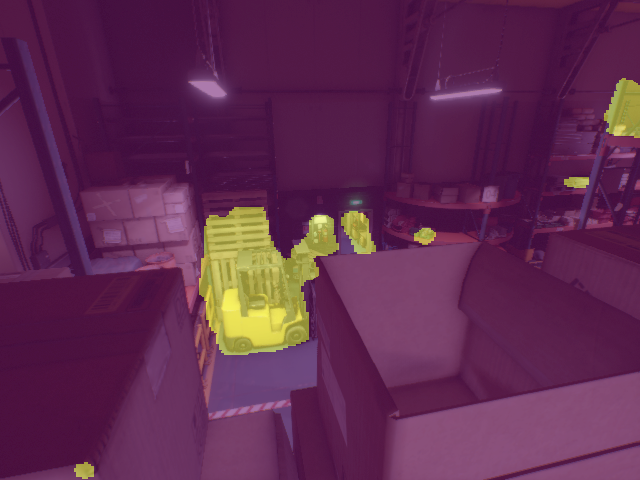} &
        \includegraphics[width=\imgwidth]{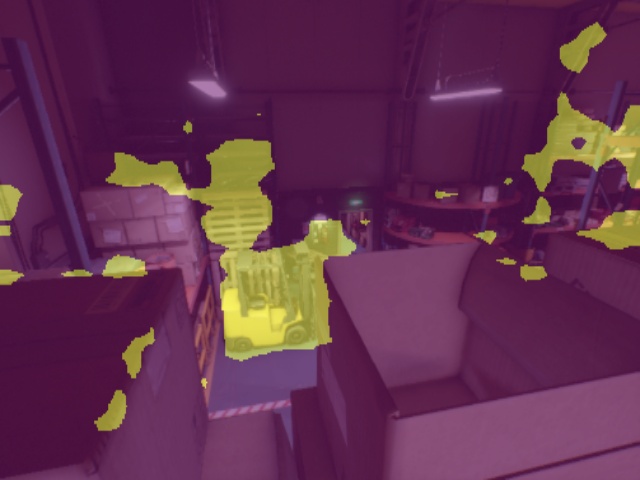} &
        \includegraphics[width=\imgwidth]{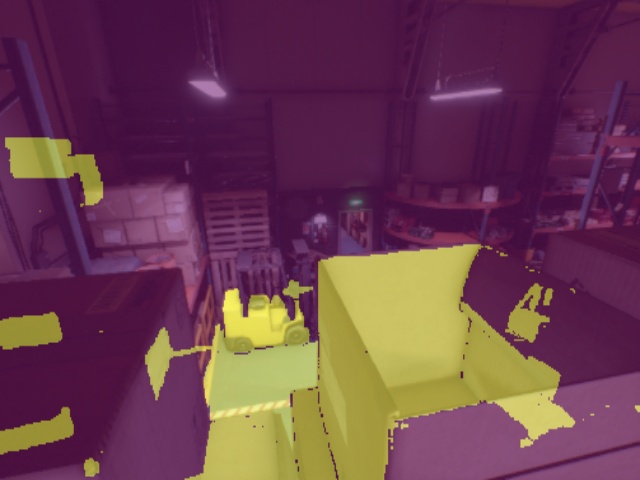} &
        \includegraphics[width=\imgwidth]{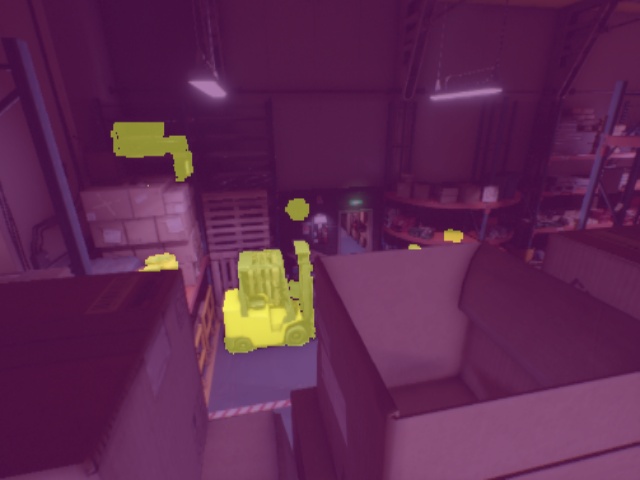} &
        \includegraphics[width=\imgwidth]{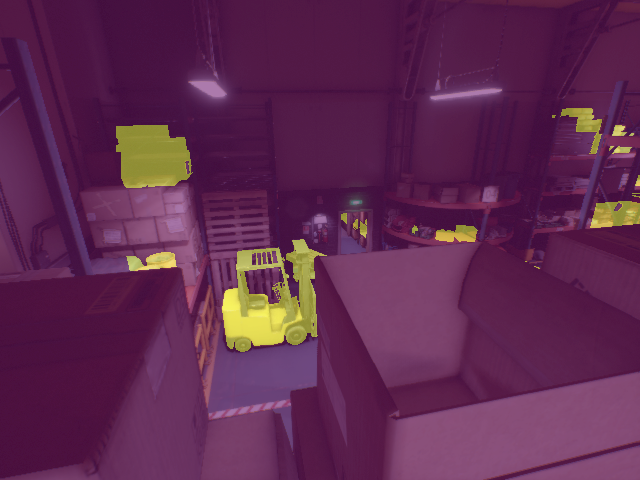}  \\

        \phantom{\includegraphics[width=\imgwidth]{figures/05_use/t0.jpg}}
        & \includegraphics[width=\imgwidth]{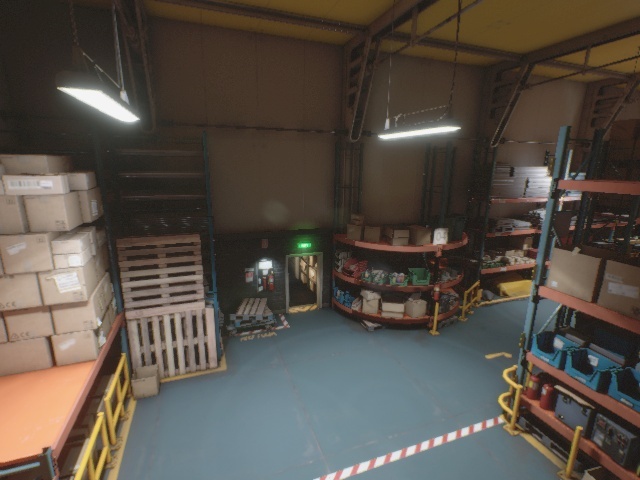} &
       \includegraphics[width=\imgwidth]{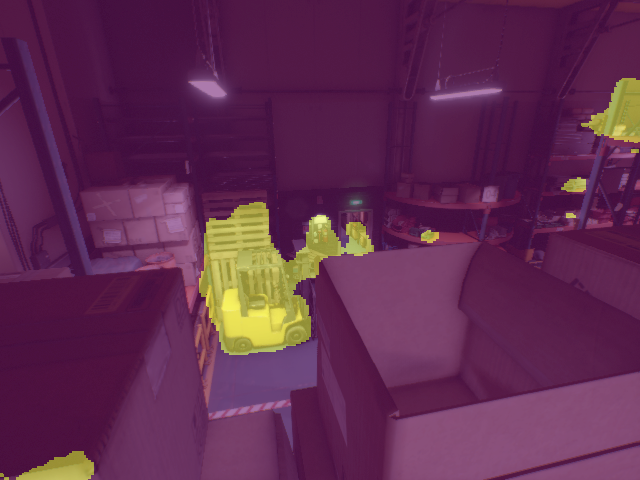} &
        \includegraphics[width=\imgwidth]{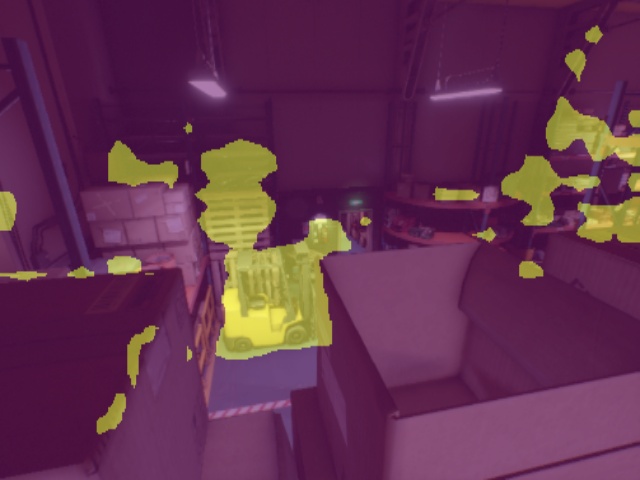} &
        \includegraphics[width=\imgwidth]{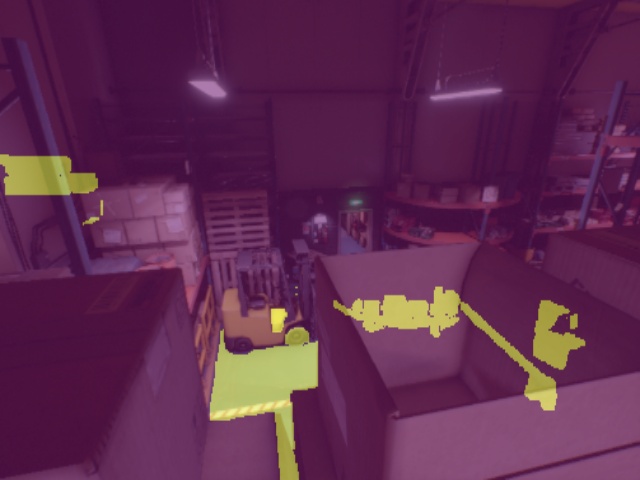} &
        \includegraphics[width=\imgwidth]{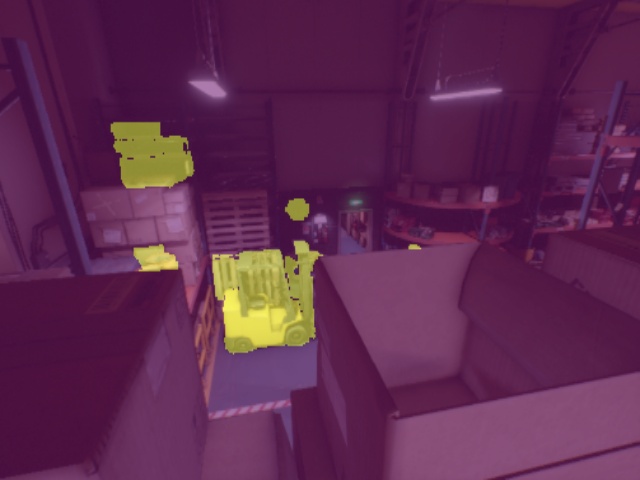} &
        \includegraphics[width=\imgwidth]{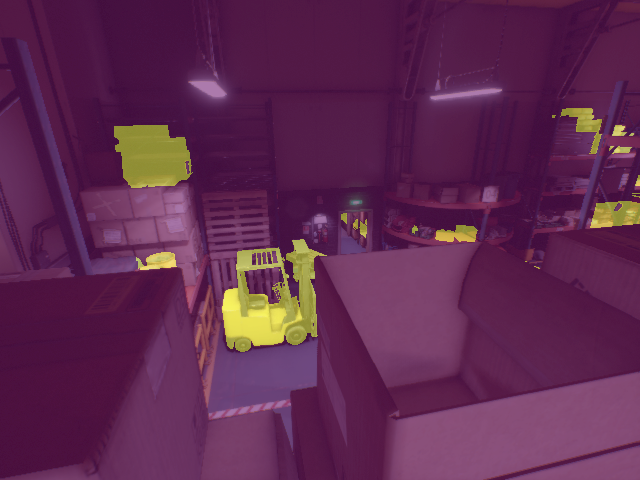}  \\

            \midrule
        % \multirow{3}{*}{\raisebox{-0.5\height}{\includegraphics[width=\imgwidth]{figures/pscd_resize/t0.jpg}}} 
        % \multirow{3}{*}{\rotatebox[origin=c]{90}{\textbf{PASLCD}}}
         % \phantom{\includegraphics[width=\pscdwidth]{figures/pscd_resize/t0.jpg}} &
       % \raisebox{-0.5\height}{\includegraphics[width=\pscdwidth]{figures/pscd_resize/t0.jpg}} &
       % \raisebox{-0.15\height}{\includegraphics[width=\pscdwidth]{figures/pscd_resize/t0.jpg}} &
     % \multirow{2}{*}{\adjustbox{valign=c}{\includegraphics[width=\pscdwidth]{figures/pscd_resize/t0.jpg}}} &
      % \phantom{\includegraphics[width=\pscdwidth]{figures/pscd_resize/t0.jpg}} &

        \multirow{2}{*}{\raisebox{-0.5\height}{\includegraphics[width=\pscdwidth]{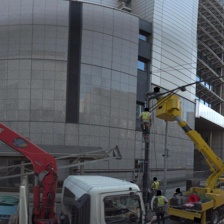}}} &
        \includegraphics[width=\pscdwidth]{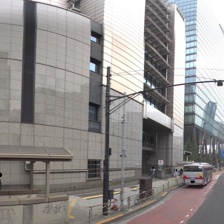} &
         \includegraphics[width=\pscdwidth]{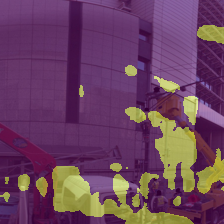} &
        \includegraphics[width=\pscdwidth]{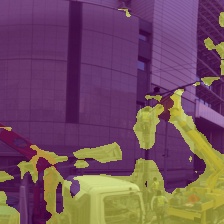} &
        \includegraphics[width=\pscdwidth]{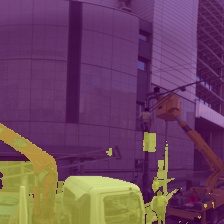} &
        \includegraphics[width=\pscdwidth]{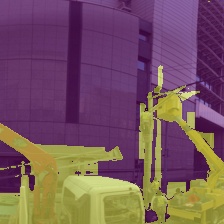} &
        \includegraphics[width=\pscdwidth]{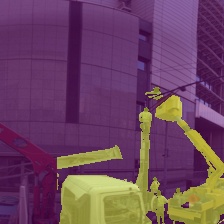}  \\

        & \includegraphics[width=\pscdwidth]{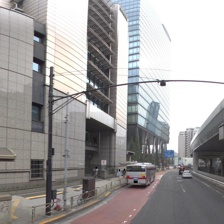} &
        \includegraphics[width=\pscdwidth]{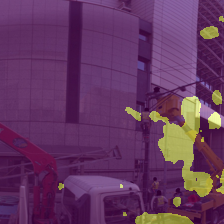} &
        \includegraphics[width=\pscdwidth]{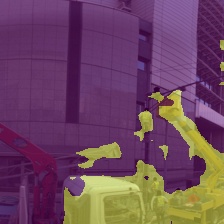} &
        \includegraphics[width=\pscdwidth]{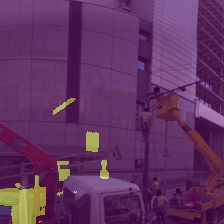} &
        \includegraphics[width=\pscdwidth]{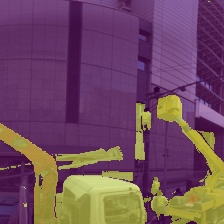} &
        \includegraphics[width=\pscdwidth]{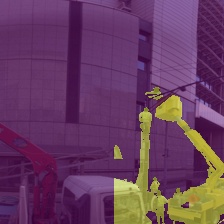}  \\

        % \phantom{\includegraphics[width=\pscdwidth]{figures/pscd_resize/t0.jpg}} &
        % \includegraphics[width=\pscdwidth]{figures/pscd_resize/stride3_t1.jpg} &
        % \includegraphics[width=\pscdwidth]{figures/cscdnet_pscd/mask3_vis.png} &
        % \includegraphics[width=\pscdwidth]{figures/pscd_resize/rscd3_mask_vis.jpg} &
        % \includegraphics[width=\pscdwidth]{figures/pscd_resize/base3_mask_vis.jpg} &
        % \includegraphics[width=\pscdwidth]{figures/pscd_resize/stride3_mask_vis.jpg} &
        % \includegraphics[width=\pscdwidth]{figures/pscd_resize/gt3_vis.jpg}  \\

        \bottomrule
    \end{tabular}
    \caption{Visualization of our results compared with baseline methods. From top to bottom, the three groups correspond to PASLCD$^\dagger$, ChangeSim$^\dagger$ (stride-5/10/15), and PSCD$^\dagger$ (stride-1/2). Our method produces cleaner masks and handles viewpoint changes more robustly.}
    % \vspace{1em}
    \label{fig:qualitative_results}
    % \vspace{-1em}
\end{figure*}

To evaluate SCD methods under unaligned settings, previous work~\cite{rbcd} preprocess VL-CMU-CD~\cite{vl_cmu_cd} and PSCD~\cite{pscd} into viewpoint-varied versions. However, VL-CMU-CD suffers from constrained viewpoint variations, along with noisy ground truth~\cite{c3po} and degraded image quality due to artifact introduced by image augmentation. Therefore, we did not adopt this dataset in our experiments. For PSCD, we follow  RSCD~\cite{rbcd} to crop panoramic images into 15 overlapping views, and generate two misalignment subsets with neighboring view intervals of 1 and 2. However, this setup introduces only two fixed rotational camera changes without translation. This limited variability fails to adequately assess model performance under diverse viewpoints. Moreover, the dataset exclusively contains street-level autonomous driving scenarios, which fails to capture the diversity of real-world environments. To enable a more comprehensive evaluation, we introduce two new datasets ChangeSim~\cite{changesim} and PASLCD~\cite{mv3dcd} into the unaligned SCD evaluation. The properties of all evaluation datasets are summarized in Table~\ref{tab:benchmark-summary}. We provide details of the newly introduced datasets and preprocessing procedures below.

ChangeSim is a simulated dataset of indoor warehouse scenes, comprising 10 scenes (6 for training and 4 for testing) with 80 sequences, rendered along simulated UAV trajectories with diverse change types and pixel-level annotations. We construct three unaligned subsets by sampling neighboring view intervals of 5, 10, and 15 from the coarsely aligned sequences each representing a different degree of viewpoint difference. For each sampled pair, we compute the visual overlap regions using the provided depth and camera poses and mask out the ground-truth labels outside the overlap.

PASLCD is a recently proposed challenging 3D change detection dataset comprising 10 real-world scenes, including 5 indoor and 5 outdoor. The dataset covers diverse change types, with approximately 70\% structural changes and 30\% surface-level modifications. For each scene, multi-view images are provided with SfM-estimated~\cite{colmap} camera poses, and three subsets are constructed by pairing each test image with its 5, 10, and 15 nearest reference images. As PASLCD does not provide per-view depth maps, we first reconstruct each scene using 3DGS~\cite{3dgs} based on the camera poses and multi-view images. Depth maps for image pairs are then rendered from 3DGS, from which we compute the overlap regions and process the ground truth masks consistent with the processing applied to ChangeSim.

\subsection{Metrics}
Following prior work, we use F1-score, the harmonic mean of precision and recall, as the main evaluation metric, and report mIoU in the ablation study.

\begin{table*}[!ht]
    \setlength{\tabcolsep}{3pt}
    \centering
    \caption{Quantitative comparison of training-based and training-free scene change detection (SCD) methods on the ChangeSim, PSCD, and PASLCD datasets under varying disparity conditions. The best score in each column is highlighted in \textbf{bold}, and the second-best score is indicated with \underline{underline}. We report the F1-score, higher values indicate better performance.}
    \resizebox{\textwidth}{!}{
        \begin{tabular}{l | l | c | c c c c c | c c c c | c c c c}
            \toprule
            \multicolumn{3}{c|}{} &
            \multicolumn{5}{c|}{ChangeSim$^\dagger$~\cite{changesim}} &
            \multicolumn{4}{c|}{PSCD$^\dagger$~\cite{pscd}} &
            \multicolumn{4}{c}{PASLCD$^\dagger$~\cite{mv3dcd}} \\
            
            Training Dataset & Method & Aug & Aligned & stride-5 & stride-10 & stride-15 & Avg & Aligned & stride-1 & stride-2 & Avg & stride-5 & stride-10 & stride-15 & Avg \\
            
            \midrule
            \multirow{6}{*}{\textbf{ChangeSim}} 
                & CSCDNet~\cite{pscd}  & \xmark & 0.368 & 0.334 & 0.305 & 0.280 & 0.322& 0.285 & 0.233 & 0.193 & 0.237& 0.122 & 0.120 & 0.121 & 0.121\\
                & CSCDNet~\cite{pscd}  & \cmark & 0.354 & 0.328 & 0.305 & 0.285 & 0.318& 0.226 & 0.196 & 0.162 & 0.195& 0.103 & 0.102 & 0.103 & 0.103\\
                & C-3PO~\cite{c3po}   & \xmark  & 0.403 & 0.351 & 0.303 & 0.268 & 0.331& 0.206 & 0.137 & 0.121 & 0.155& \underline{0.131} & 0.120 & 0.128 & 0.126\\
                & C-3PO~\cite{c3po}   & \cmark  & 0.384 & 0.338 & 0.299 & 0.265 & 0.322& 0.227 & 0.134 & 0.102 & 0.154& 0.097 & 0.089 & 0.090 & 0.092\\
                & RSCD~\cite{rbcd} & \xmark  & 0.364 & 0.315 & 0.280 & 0.255 & 0.303& 0.347  & 0.277 & 0.209 & 0.278& 0.085 & 0.083 & 0.081 & 0.083\\
                & RSCD~\cite{rbcd} & \cmark  & 0.392 & 0.355 & 0.328 & 0.305 & 0.345& 0.312 & 0.276 & 0.219 & 0.269& 0.073 & 0.072 & 0.071 & 0.072\\
            \midrule
            \multirow{6}{*}{\textbf{PSCD}} 

                & CSCDNet~\cite{pscd}  & \xmark  & 0.274 & 0.238 & 0.211 & 0.189 & 0.228& \underline{0.533} & 0.395 & 0.296 & 0.408& 0.114 & 0.115 & 0.116 & 0.115\\
                & CSCDNet~\cite{pscd}  & \cmark  & 0.243 & 0.212 & 0.188 & 0.171 & 0.204& 0.503 & \underline{0.443} & \textbf{0.411} & \textbf{0.452} & 0.099 & 0.097 & 0.100 & 0.099\\
                & C-3PO~\cite{c3po}   & \xmark  & 0.303 & 0.257 & 0.220 & 0.192 & 0.243& 0.486 & 0.337 & 0.232 & 0.352& 0.076 & 0.064 & 0.071 & 0.070\\ 
                & C-3PO~\cite{c3po}   & \cmark  & 0.329 & 0.267 & 0.219 & 0.183 & 0.249& 0.458 & 0.372 & 0.298 & 0.376& 0.101 & 0.102 & 0.097 & 0.100\\ 
                 & RSCD~\cite{rbcd} & \xmark  & 0.277 & 0.235 & 0.208 & 0.187 & 0.227& \textbf{0.544} & 0.356 & 0.241 & 0.380& 0.129 & 0.128 & 0.120 & 0.126\\
                & RSCD~\cite{rbcd} & \cmark  & 0.240 & 0.206 & 0.183 & 0.166 & 0.199& 0.490 & 0.419 & \underline{0.353} & 0.421& \underline{0.131} & \underline{0.129} & \underline{0.131} & \underline{0.130}\\
            \midrule
            \multirow{3}{*}{\textbf{Zero-Shot}} 
                & ZSCD~\cite{zero} & - & 0.288 & 0.292 & 0.281 & 0.266 & 0.282 & 0.298 & 0.254 & 0.219 & 0.257 & 0.120 & 0.102 & 0.094 & 0.105 \\
                & GeSCF~\cite{gescf} & - & \underline{0.513} & \underline{0.451} & \underline{0.396} & \underline{0.346} & \underline{0.426} & 0.330 & 0.269 & 0.166 & 0.255 & 0.094 & 0.088 & 0.088 & 0.090 \\
                & \textbf{Ours}    & - & \textbf{0.569} & \textbf{0.557} & \textbf{0.517} & \textbf{0.481} & \textbf{0.531} & \underline{0.533} & \textbf{0.454} & 0.350 & \underline{0.446} & \textbf{0.351} & \textbf{0.312} & \textbf{0.291} & \textbf{0.318} \\
            \bottomrule
        \end{tabular}
    }
    \label{tab:table-backbone}
\end{table*}

\subsection{Baselines}
In our experiments, we compare our proposed framework against three representative training-based SCD methods, CSCDNet~\cite{pscd}, C-3PO~\cite{c3po}, and RSCD~\cite{rbcd}, as well as two training-free approaches, ZSCD~\cite{zero} and GeSCF~\cite{gescf}. In addition, for the ablation study, we construct a baseline by integrating the SOTA optical flow method SEA-RAFT~\cite{sea-raft} with our Geometry-guided Change Mask Prediction module, replacing geometric correspondences with optical flow and without handling occlusions. Among them, CSCDNet~\cite{pscd} and RSCD~\cite{rbcd} explicitly incorporate mechanisms to handle viewpoint misalignment. We retrain all three training-based baselines on ChangeSim~\cite{changesim} and PSCD~\cite{pscd}.
To further examine the generalization of training-based methods to varying viewpoint differences, we adopt the viewpoint augmentation strategy introduced in RSCD~\cite{rbcd} and design fine-grained experimental groups. Specifically, for each method, we evaluate two variants: one without augmentation and one with a hybrid multi-viewpoint augmentation strategy. On ChangeSim, aligned pairs are sampled with probability 0.4 and unaligned pairs at three viewpoint intervals (stride-5/10/15) with probability 0.2 each. On PSCD, aligned pairs are sampled with probability 0.6 and unaligned pairs at two viewpoint intervals (stride-1/2) with probability 0.2 each. In addition, on ChangeSim, we introduce a single-viewpoint augmentation setting, where training uses only one viewpoint interval of unaligned pairs, to assess model robustness under unseen viewpoint conditions. All experiments are conducted on NVIDIA V100 GPUs with a batch size of 4, using the default settings of the respective methods.

\section{Results}

\subsection{Viewpoint Robustness and Cross-Dataset Generalization}

Table~\ref{tab:table-backbone} presents the F1-scores of baselines and proposed method across different datasets, while also analyzing the effect of different viewpoint augmentation on training-based models. For the augmentation group, we adopt the aforementioned hybrid augmentation strategy. Furthermore, we evaluate the cross-dataset generalization of these methods. The key observations are summarized as follows:

\textbf{1) Robustness to Viewpoint Variations:} 
Our framework demonstrates strong performance across different viewpoint intervals and datasets, highlighting its robustness to viewpoint variations. On ChangeSim, our method achieves an average F1-score of 0.531, substantially outperforming all baselines and surpassing the second-best method (0.426) by 24.6\%. On PASLCD, it consistently achieves the best results across all viewpoint variation levels, with an average score of 0.318 compared to the second-best method (0.130). On PSCD, our method achieves competitive performance across viewpoint separations, obtaining the second-highest average score. This is partly due to PSCD’s cropped panorama pairs with fixed overlaps, which favor some training-based methods that implicitly exploit such regularities. In contrast, our framework relies on general geometric reasoning rather than dataset-specific patterns, ensuring robust performance across datasets.

\textbf{2) Different-Viewpoint Augmentation Does Not Always Help:}
Table~\ref{tab:table-backbone} shows that different-viewpoint augmentation does not consistently improve training-based methods. RSCD benefits from augmentation in the unaligned setting due to its full-image cross-attention, which provides stronger cross-view matching. This also explains why RSCD improves even on the aligned ChangeSim setting, where residual misalignments remain. In contrast, C-3PO and CSCDNet exhibit mixed behavior: augmentation improves their performance on PSCD but reduces it on ChangeSim. Overall, across nearly all methods, augmentation tends to cause performance drops under the aligned setting.

\textbf{3) Strong Cross-dataset Generalization of Our Method:} 
The results in Table~\ref{tab:table-backbone} show that training-based methods suffer substantial performance drops under zero-shot cross-dataset evaluation, and both training-based and training-free baselines struggle on the challenging real-world PASLCD dataset. In contrast, our method maintains consistently strong performance, demonstrating robust cross-dataset generalization.

\subsection{Qualitative Comparison}

In Fig.~\ref{fig:qualitative_results}, we present qualitative comparisons across different datasets and viewpoint differences. For ChangeSim and PSCD, training-based models are trained on the corresponding dataset with viewpoint augmentation, while for PASLCD, we use the model trained on PSCD with viewpoint augmentation.
The results show that training-based methods produce noisy masks even on in-domain data. For instance, RSCD exhibits a considerable number of false positives on ChangeSim (group 2, 4th column). Their performance further degrades on unseen datasets like PASLCD, where CSCDNet misclassifies complex background regions as changes, and RSCD produces coarse masks despite roughly detecting the changed regions. Meanwhile, the training-free method GeSCF lacks the ability to handle viewpoint variations, resulting in substantial performance degradation under larger viewpoint differences (group 2, 5th column) and even near-complete failure in some cases (group 3, 5th column). In contrast, our method effectively handles viewpoint changes, maintaining stable performance and producing more accurate masks even under significant viewpoint shifts. Our method can also precisely detect small or challenging changed objects, such as the desktop items and background broom in group 1. Overall, our approach consistently achieves higher accuracy and stronger viewpoint robustness across datasets and varying viewpoint conditions.

\subsection{Viewpoint Generalization}

\begin{table}[t]
    \setlength{\tabcolsep}{3pt}
    \centering
    \caption{F1-score of training-based models under different training strategies on ChangeSim. Each group represents models trained either without augmentation or with a single-viewpoint interval or with a hybrid augmentation strategy.}
    \resizebox{0.48\textwidth}{!}{
        \begin{tabular}{l | l | c c c c}
            \toprule
            Augmentation & Method & Aligned & stride-5 & stride-10 & stride-15 \\
            \midrule
            \multirow{3}{*}{w/o Aug} 
                & RSCD~\cite{rbcd} & 0.364 & 0.315 & 0.280 & 0.255 \\
                & C-3PO~\cite{c3po} & \underline{0.403} & 0.351 & 0.303 & 0.268 \\
                & CSCDNet~\cite{pscd} & 0.368 & 0.334 & 0.305 & 0.280 \\
            \midrule
            \multirow{3}{*}{stride-5} 
                & RSCD~\cite{rbcd} & 0.398 & \underline{0.356} & 0.324 & 0.297 \\
                & C-3PO~\cite{c3po} & 0.396 & 0.343 & 0.301 & 0.265 \\
                & CSCDNet~\cite{pscd}   & 0.367 & 0.340 & 0.316 & 0.293 \\
            \midrule
            \multirow{3}{*}{stride-10} 
                & RSCD~\cite{rbcd} & 0.388 & 0.345 & 0.315 & 0.291 \\
                & C-3PO~\cite{c3po}   & 0.374 & 0.328 & 0.291 & 0.260 \\
                & CSCDNet~\cite{pscd}   & 0.333 & 0.313 & 0.294 & 0.275 \\
            \midrule
            \multirow{3}{*}{stride-15} 
                & RSCD~\cite{rbcd} & 0.383 & 0.339 & 0.311 & 0.288 \\
                & C-3PO~\cite{c3po}   & 0.380 & 0.331 & 0.291 & 0.260 \\
                & CSCDNet~\cite{pscd}   & 0.324 & 0.302 & 0.286 & 0.2713 \\
            \midrule
            \multirow{3}{*}{Hybrid} 
                & RSCD~\cite{rbcd} & 0.392 & 0.355 & \underline{0.328} & \underline{0.305} \\
                & C-3PO~\cite{c3po} & 0.384 & 0.338 & 0.299 & 0.265 \\
                & CSCDNet~\cite{pscd} & 0.354 & 0.328 & 0.305 & 0.285 \\
            \midrule
            w/o Aug & Ours &  \textbf{0.569}  & \textbf{0.557} & \textbf{0.517} & \textbf{0.481} \\
            \bottomrule
        \end{tabular}
    }
    \label{tab:changesim-augmentation}
\end{table}

Beyond evaluating the effect of different viewpoint augmentation strategies on training-based methods, we further examine the single-viewpoint augmentation setting on ChangeSim to assess their generalization ability to unseen viewpoint differences. As shown in Table~\ref{tab:changesim-augmentation}, training-based models exhibit very limited viewpoint generalization: when trained with only stride-10 or stride-15 pairs, their performance drops significantly on aligned cases and on smaller viewpoint intervals (stride-5). This indicates that these models fail to learn transferable geometric knowledge from limited single-viewpoint samples. More importantly, even under the corresponding large-viewpoint conditions, single-viewpoint augmentation does not yield a clear advantage over small-viewpoint augmentation. Overall, these results highlight the inherent limitations of training-based approaches in handling multi-view variations. In contrast, our method consistently achieves stable and superior performance across different viewpoint levels without requiring any augmentation.

\subsection{Ablation Study}

We conducted an ablation study on the main modules of our method using the PSCD and PASLCD datasets, evaluating three settings. The first setting (OF + FC) establishes dense pixel correspondences using the optical flow method SEA-RAFT~\cite{sea-raft}, which are then directly used to guide feature correlation and mask fusion. The second setting (GM + FC) replaces optical flow with the proposed geometric correspondences, while filtering out pixels outside the overlapping region. The third setting (GM + FC + Occ) further incorporates occlusion handling. For evaluation, PSCD~\cite{pscd} metrics are reported as the average over stride-1 and stride-2, while PASLCD~\cite{mv3dcd} metrics are averaged over stride-5, stride-10, and stride-15.

\begin{table}[ht]
\setlength{\tabcolsep}{6pt} 
\centering
\small 
\caption{Ablation study on the PSCD and PASLCD datasets.}
\begin{tabular}{l | c c | c c }
    \toprule
    \multirow{2}{*}{\textbf{Method}} & \multicolumn{2}{c}{PSCD$^\dagger$~\cite{pscd}} & \multicolumn{2}{c}{PASLCD$^\dagger$~\cite{mv3dcd}}  \\
    & F1-Score & mIoU & F1-Score & mIoU  \\
    \midrule
    OF + FC & 0.244 & 0.155 & 0.094 & 0.052 \\
    GM + FC & 0.391 & 0.274 & 0.237 & 0.148 \\
    GM + FC + Occ & \textbf{0.402} & \textbf{0.282} & \textbf{0.321} & \textbf{0.218} \\
    \bottomrule
\end{tabular}
\label{tab:ablation}
\end{table}

As shown in Table~\ref{tab:ablation}, leveraging geometric correspondences instead of optical flow leads to a significant performance boost, demonstrating that geometric priors enable more reliable identification of visual overlaps and robust correspondence establishment for unaligned SCD. Incorporating occlusion handling further enhances performance, particularly on real-world multi-view datasets PASLCD, where the F1-score improves by 0.084 (35\%), underscoring the importance of occlusion reasoning in practical scenarios.

\section{Conclusion}

In this work, we tackled the challenging problem of scene change detection under unaligned viewpoints by incorporating geometric priors into SCD for the first time. Leveraging GFM, we construct robust geometric correspondences within visual overlap and detect occlusions, which are then integrated with SAM to form a training-free, geometry-aware framework that delivers reliable performance across diverse viewpoint differences and scenarios. 
Superior results of our approach demonstrate the critical role of geometric priors in enabling more robust and generalizable SCD. Moreover, by relaxing the conventional assumption of viewpoint alignment, our approach further broadens the applicability of SCD to more flexible and realistic scenarios.

\bibliographystyle{IEEEtran}
\bibliography{reference}

\end{document}